\documentclass[conference]{IEEEtran}
\IEEEoverridecommandlockouts
\renewcommand{\thesection}{\Roman{section}}
\usepackage{booktabs}
\usepackage{fancyhdr}
\usepackage{algorithm}
\usepackage{fancyhdr}
\usepackage{algorithmic}

\usepackage{caption}
\usepackage{graphicx}
\usepackage{array}
\usepackage{adjustbox}
\usepackage{hyperref}
\usepackage{float}

\usepackage{cite}
\usepackage{multirow}
\usepackage{amsmath,amssymb,amsfonts}
\usepackage{algorithmic}
\usepackage{graphicx}
\usepackage{textcomp}
\usepackage{comment}

\def\BibTeX{{\rm B\kern-.05em{\sc i\kern-.025em b}\kern-.08em
    T\kern-.1667em\lower.7ex\hbox{E}\kern-.125emX}}
    
\fancypagestyle{firstpage}{
  \fancyhf{}
  \fancyhead[L]{\fontsize{8}{9}\selectfont
  2026 IEEE 2nd International Conference on Quantum Photonics, Artificial Intelligence, and Networking (QPAIN)\\
  16 -- 18 April 2026, Chittagong, Bangladesh}
  \fancyfoot[L]{\fontsize{8}{9}\selectfont
  979-8-3315-4990-9/26/\$31.00 \copyright2026 IEEE}

}

\title{Adaptive Enhancement and Dual-Pooling Sequential Attention for Lightweight Underwater Object Detection with YOLOv10}
\author
{\IEEEauthorblockN{Md. Mushibur Rahman\textsuperscript{1}, Umme Fawzia Rahim\textsuperscript{2}, Enam Ahmed Taufik\textsuperscript{3}}
\IEEEauthorblockA{\textsuperscript{1,2}Department of Computer Science and Engineering, Dhaka University of Engineering \& Technology\\ Gazipur, 1707, Bangladesh\\
\textsuperscript{3}Department of Computer Science and Engineering, BRAC University\\
Dhaka, Bangladesh \\
\textsuperscript{1}mushiburrahman5@gmail.com, \textsuperscript{2}fawzia.rahim@duet.ac.bd, 
\textsuperscript{3}enam.ahmed.taufik@gmail.com
}}

\begin{document}
\maketitle
\thispagestyle{firstpage}
\pagestyle{plain}

\begin{abstract}
Underwater object detection constitutes a pivotal endeavor within the realms of marine surveillance and autonomous underwater systems; however, it presents significant challenges due to pronounced visual impairments arising from phenomena such as light absorption, scattering, and diminished contrast. In response to these formidable challenges, this manuscript introduces a streamlined yet robust framework for underwater object detection, grounded in the YOLOv10 architecture. The proposed method integrates a Multi-Stage Adaptive Enhancement module to improve image quality, a Dual-Pooling Sequential Attention (DPSA) mechanism embedded into the backbone to strengthen multi-scale feature representation, and a Focal Generalized IoU Objectness (FGIoU) loss to jointly improve localization accuracy and objectness prediction under class imbalance. Comprehensive experimental evaluations conducted on the RUOD and DUO benchmark datasets substantiate that the proposed DPSA\_FGIoU\_YOLOv10n attains exceptional performance, achieving mean Average Precision (mAP) scores of 88.9\% and 88.0\% at IoU threshold 0.5, respectively. In comparison to the baseline YOLOv10n, this represents enhancements of 6.7\% for RUOD and 6.2\% for DUO, all while preserving a compact model architecture comprising merely 2.8M parameters. These findings validate that the proposed framework establishes an efficacious equilibrium among accuracy, robustness, and real-time operational efficiency, making it suitable for deployment in resource-constrained underwater settings.

\end{abstract}


\section{Introduction}

Underwater object detection (UOD) is a foundational technology for marine ecological monitoring, autonomous subsea navigation, and intelligent resource management. However, underwater visual perception is inherently constrained by wavelength dependent absorption, scattering, and non-uniform illumination \cite{Akkaynak2018}. These optical phenomena induce color distortion, severe contrast degradation, and boundary blurring, which significantly compromise the reliability of detection models originally developed for terrestrial environments. The proliferation of deep learning has established convolutional neural network (CNN) based detectors as the standard paradigm for real-time visual perception. These architectures are deployed on resource constrained platforms, such as autonomous underwater vehicles (AUVs) and remotely operated vehicles (ROVs). This degradation stems from early-stage feature extraction failure due to poor image quality \cite{Li2017Survey}, as well as feature aggregation strategies and loss functions that do not adequately account for class imbalance and localization uncertainty.

Recent research  efforts have sought to mitigate these challenges by integrating image enhancement, attention-based refinement, and specialized optimization objectives. Deterministic enhancement methods restore color fidelity and contrast, thereby stabilizing feature representations \cite{islam2020fast}. Furthermore, attention mechanisms improve robustness by prioritizing salient targets while suppressing environmental noise \cite{Woo2018CBAM}. Advanced loss formulations, including Intersection-over-Union (IoU) variants and focal-weighted objectives, have also demonstrated utility in improving bounding box regression and mitigating foreground-background imbalance \cite{Rezatofighi2019GIoU}. However, many existing methods rely on computationally intensive enhancement or heavy attention modules, limiting real-time deployment. The systematic integration of deterministic preprocessing, lightweight attention, and robust loss optimization remains underexplored.

In this work, we propose a lightweight and robust underwater object detection framework designed to enhance detection accuracy while preserving real-time efficiency. The primary contributions of this study can be summarized as follows:

\begin{itemize}
    \item Introduce a Multi-Stage Adaptive Enhancement pipeline to correct color distortion, enhance contrast, and preserve structural details in degraded underwater images.
    \item We introduce a lightweight Dual-Pooling Sequential Attention mechanism that applies sequential channel and spatial attention to strengthen small object feature representation and suppress complex underwater backgrounds.
    \item We develop a hybrid loss function that jointly addresses class imbalance, improves bounding box localization, and enhances objectness calibration by integrating Focal Loss, Generalized IoU Loss, and Objectness Focal Loss.
    \item Comprehensive experiments on standard underwater benchmarks demonstrate consistent mAP improvements while maintaining real-time inference performance suitable for embedded deployment.
\end{itemize}

%The remainder of this paper is organized as follows. Section II presents the literature review, Section III details the datasets, methodology, and experimental setup, Section IV discusses the results, and Section V concludes the paper with directions for future research.

\section{Related Work}
The evolution of underwater object detection (UOD) is characterized by sustained efforts to address severe optical degradation, including wavelength-dependent absorption, backscattering, and the presence of small, densely packed targets. In Recent years, research has shifted toward end-to-end architectures optimized for real-time performance on resource-constrained hardware. Initial research focused on lightweight adaptations of the YOLO framework. For example, YOLO-UOD \cite{p6} adopted YOLOv7 as the baseline detector and achieved strong underwater detection performance on the URPC2018 dataset, reporting an mAP of 83.1\% and an AP of 48.5\% with 37.3M parameters, while maintaining efficient computational complexity. Along similar lines, CEH-YOLO \cite{P7} incorporated high-order deformable attention into YOLOv8, achieving 88.4\% mAP on the DUO dataset and a detection speed of 156 FPS with only 4.4 M parameters. Similarly, Dynamic YOLO \cite{P10} enhanced detection sensitivity through deformable convolution v3 and achieved an mAP@.5 of 86.7\% on the DUO dataset with 8.2 M parameters. These principles extend to sonar imagery, For instance, UWS-YOLO \cite{P11} achieved 87.1\% mAP at 158 FPS by utilizing dynamic snake convolutions to model elongated structures. In Addition, SU-YOLO \cite{P12} introduced Spiking Neural Networks to achieve 78.8\% mAP with an energy consumption of 2.98 mJ, demonstrating a path toward ultra-low power underwater sensing. Collectively, these advancements indicate a trajectory toward high-precision frameworks that balance multi-scale feature fusion with low computational complexity.

In addition to the YOLO architecture, alternative non-YOLO single-stage detectors have been explored for the purpose of underwater object detection. Building on SSD-based designs, jiang and Wang~\cite{pp12} incorporated shallow detection layers, weighted confidence loss, and MSRCR-based image enhancement, with an mAP of 66.9\% on URPC2018, outperforming SSD, YOLOv3, and Faster R-CNN. Moving toward more advance one stage frameworks, Jia et al.~\cite{pp13} introduced an EfficientDet-based detector with enhanced channel interaction and multi-scale feature fusion, achieving 91.67\% mAP on URPC and 92.81\% mAP on a Kaggle dataset at 37.5 FPS. Similarly, Jain et al.~\cite{pp14} proposed DeepSeaNet, a modified EfficientDet model with BiSkFPN and multi-focal loss, attaining 98.63\% mAP on the Brackish dataset. More recently, Jyothimurugan et al.~\cite{pp15} developed a DCGAN augmented YOLOv6 framework for Crown-of-Thorns Starfish detection on the CSIRO dataset, achieving 93.8\% mAP@.5 with strong precision and recall while enabling real-time embedded deployment.

\section{Methodology}

\subsection{Dataset}
\begin{figure}[htbp]
    \centering
    \includegraphics[width=.85\linewidth, height=.12\textheight]{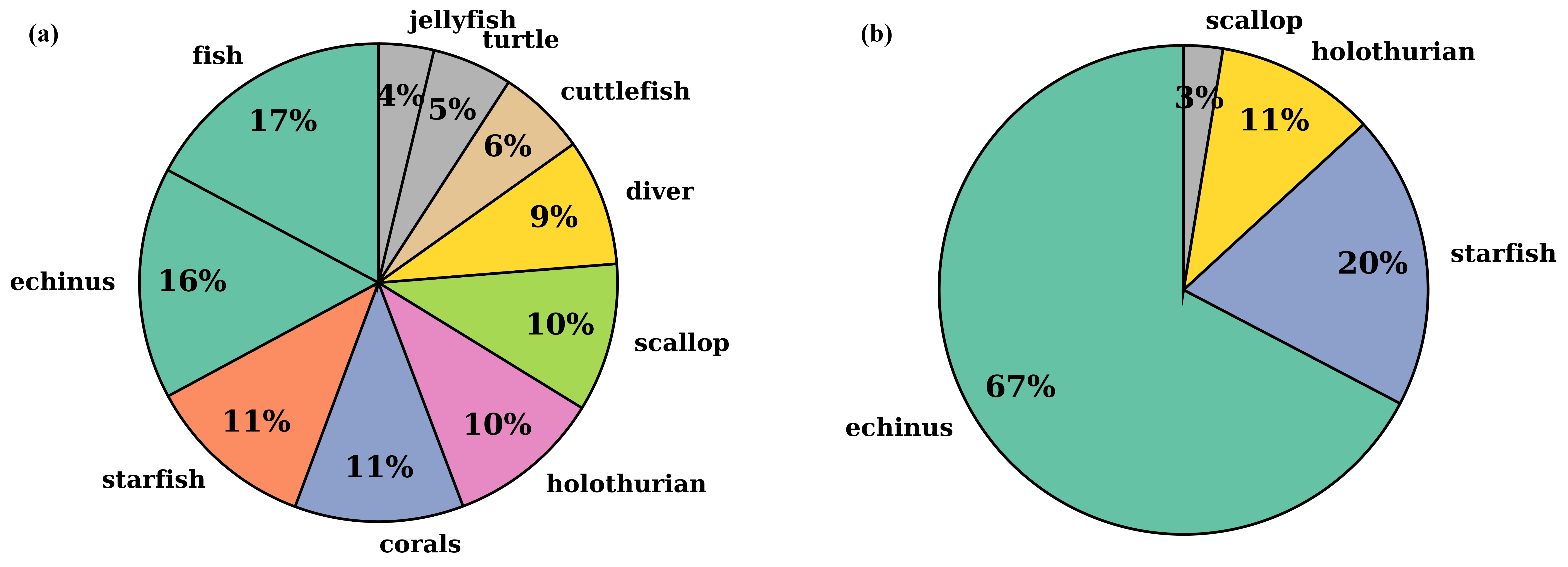}
    \caption{Distribution of samples across object categories in the datasets: (a) RUOD dataset and (b) DUO dataset.}
    \label{fig:dataset_pie_chart}
\end{figure}

To enable fair comparison with existing methods, this study uses two established benchmark datasets for underwater object detection: the RUOD dataset~\cite{d1} and the DUO dataset~\cite{d2}. The diversity in object categories, image resolutions, and class distributions makes these datasets suitable for evaluating robustness under challenging underwater conditions. The RUOD dataset consists of 9,340 images across 10 categories: fish, jellyfish, turtle, cuttlefish, diver, scallop, holothurian, coral, starfish, and echinus. All images are of size $640 \times 640$ pixels. The DUO dataset encompasses a total of 7,782 images across four categories: sea urchin, starfish, sea cucumber, and scallop. This dataset is characterized by images with five different resolutions: $3840 \times 2160$, $1920 \times 1080$, $720 \times 405$, $586 \times 480$, and $704 \times 576$. The distribution of species categories within the two datasets are shown in Figure~\ref{fig:dataset_pie_chart}. RUOD and DUO are established benchmarks that cover diverse underwater object categories and imaging conditions.

%The dataset used in this study including following: the RUOD dataset \cite{d1}, and DUO Dataset \cite{d2}. The RUOD dataset consists of 9,340 images across 10 categories: fish, jellyfish, turtle, cuttlefish, diver, scallop, holothurian, coral, starfish, and echinus. All images are 640 × 640 pixels. The DUO dataset contains 7782 images across four categories: sea urchin, starfish, sea cucumber, and scallop. This dataset contains images of the following five different resolutions: 3840 × 2160, 1920 × 1080, 720 × 405, 586 × 480 and 704 × 576. The quantity distributions of species categories in the two datasets are shown in Figure~{\ref{fig:dataset_pie_chart}}. This two datasets are established benchmarks for underwater object detection and enable fair comparison with existing methods.

\subsection{Multi-Stage Adaptive Preprocessing for Underwater Visual Perception}
\begin{figure}[htbp]
    \centering
    \includegraphics[width=.95\linewidth, height=.26\textheight]{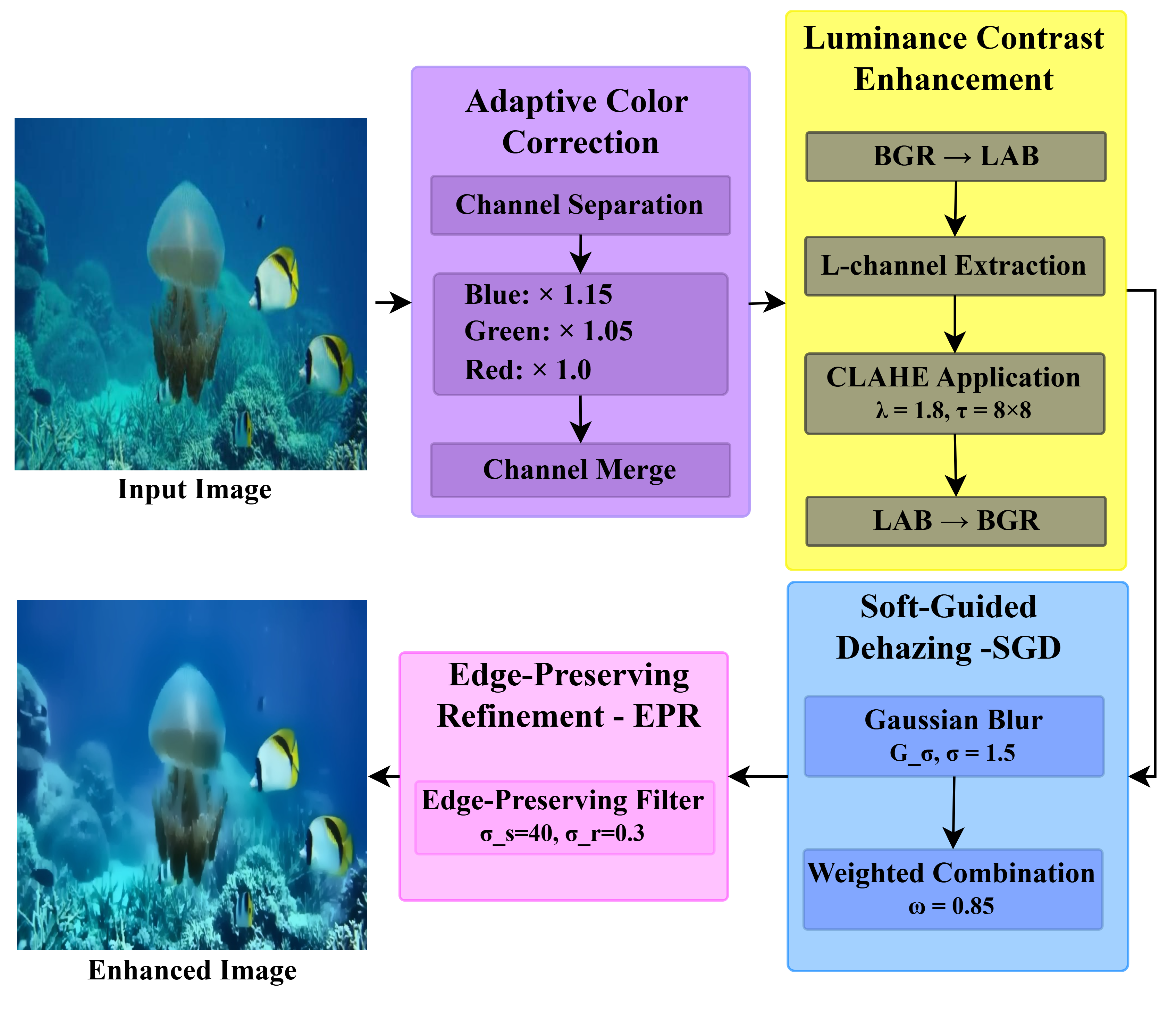}
    \caption{Structure of MAE-UVP enhancement module.}
    \label{fig:mae_uvp_arch}
\end{figure}

Underwater imaging is severely affected by wavelength-dependent attenuation and scattering, leading to color distortion, uneven illumination, and reduced contrast. In response to this issue, we proposed a Multi-Stage Adaptive Enhancement for Underwater Visual Perception (MAE-UVP) module implemented as a deterministic preprocessing framework. In Figure~\ref{fig:mae_uvp_arch}, the MAE-UVP comprises four sequential correction phases. (1) Adaptive color correction compensates for the dominant cyan bias through channel wise scaling to recover attenuated red components. (2) Luminance contrast Enhancement transitions the image into CIELAB color space and applies CLAHE exclusively to the luminance channel, thereby augmenting local contrast without introducing color distortion. (3) Soft-Guided Dehazing (SGD), employs a Gaussian-guided prior to delicately attenuate forward scattering haze while maintaining edge clarity and preventing halo artifacts. (4) Edge Preserving Refinement applies edge aware filtering to enhance object boundaries and reduce noise in homogeneous regions. MAE-UVP contains no learnable parameters, all hyperparameters are empirically determined and uses fixed, non-learnable hyperparameters to ensure deterministic, low-overhead, and reproducible enhancement.
\begin{figure*}[htbp]
    \centering
    \includegraphics[width=.95\linewidth, height=.6\textheight]{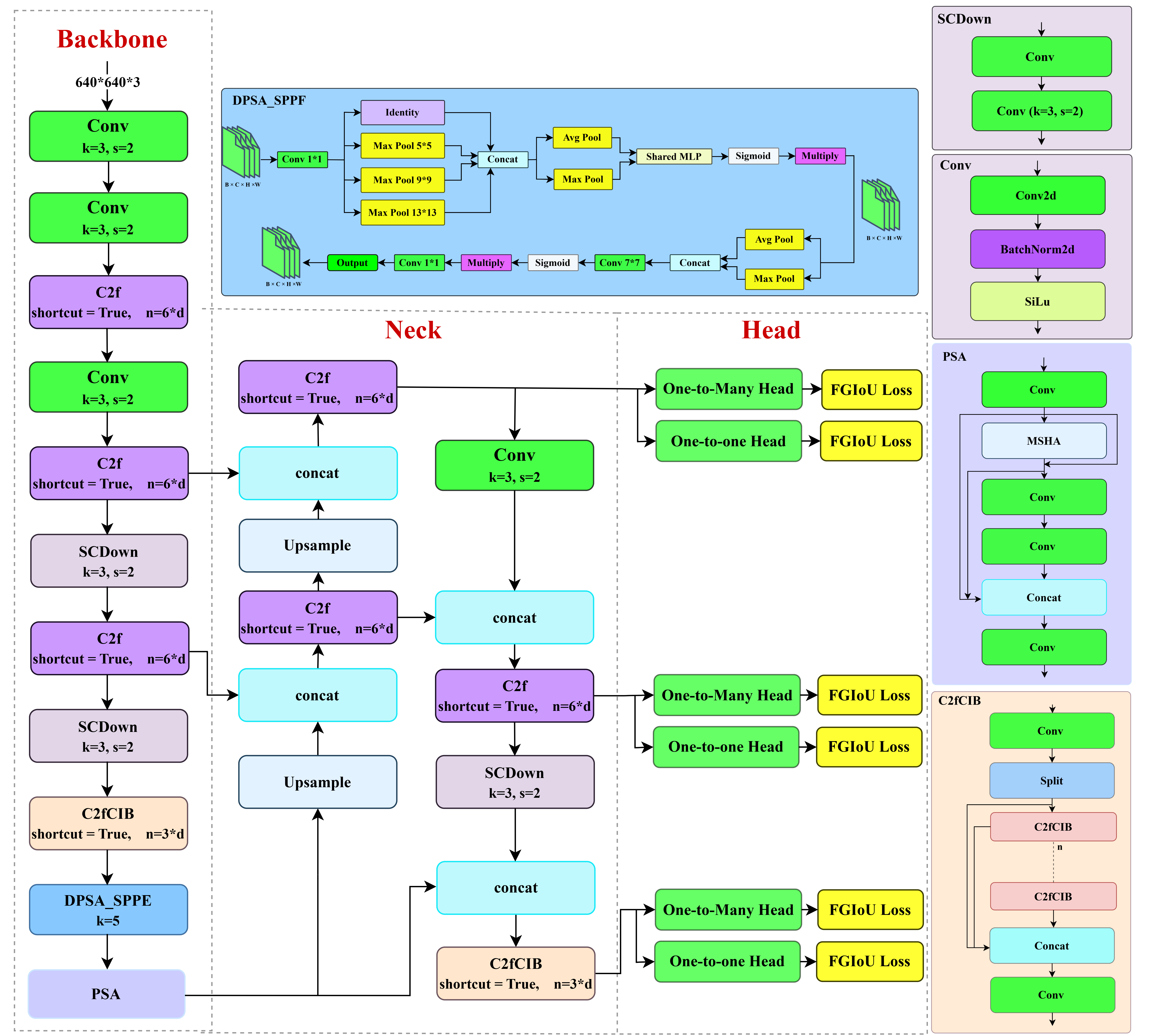}
   \caption{Proposed YOLOv10 architecture with SPPF replaced by DPSA\_SPPF in the backbone and modified detection head.}
    \label{fig:YOLOv10}
\end{figure*}
\subsection{Improved YOLOv10 Model Architecture}
The proposed architectural builds upon the foundational YOLOv10~\cite{yolov10} framework by meticulously integrating a series of critical enhancements that are both innovative and functionally impactful. Specifically, the Spatial Pyramid Pooling Fast (SPPF) layer in the backbone is replaced with the proposed DPSA\_SPPF module, which applies sequential channel and spatial attention to multi scale pooled features to enhance feature discrimination underwater conditions. This modification emphasizes salient object regions while suppressing background artifacts, without altering the backbone topology. In addition, the detection head is optimized by incorporating the proposed loss function, improving localization accuracy and objectness calibration. Notably, the PAN-FPN neck of YOLOv10~\cite{yolov10} remains unchanged, preserving the original feature aggregation strategy and ensuring fair comparison with the baseline.As shown in Figure~\ref{fig:YOLOv10}, the architecture diagram highlights the backbone and head level modifications with the end-to-end, NMS-free detection pipeline.

\subsection{Dual Pooling Sequential Attention (DPSA)}
In the current investigation, the DPSA mechanism was judiciously integrated atop the SPPF layer of the backbone network, a design decision informed by two principal considerations. Firstly, this placement allows the multi-scale features generated by SPPF to sequential attention refinement, which helps suppress underwater noise and enhance object relevant representations. Secondly, computational efficiency is achieved by processing features post concatenation, capitalizing on DPSA streamlined structure while preventing redundant attention computations across scales.

%\begin{figure}[htbp]
   % \centering
   % \includegraphics[width=.9\linewidth, height=.12\textheight]{image/DPSA.png}
   % \caption{DPSA attention mechanism.}
   % \label{fig:DPSA_SPPF}
%\end{figure}
The DPSA module consists of two sequential attention components with domain specific alterations. Channel Attention utilizes dual adaptive pooling processed through a shared two layer convolutional MLP with a fixed reduction ratio of sixteen, producing channel weights via sigmoid. Spatial Attention computes channel wise mean and maximum statistics, concatenates these, and applies a fixed 7×7 convolutional kernal with padding. The comprehensive DPSA processes features through channel attention followed by spatial attention. The DPSA\_SPPF variant reduces channels via 1×1 convolution, implements three parallel max pooling operations with kernels of 5×5, 9×9, and 13×13, concatenates multi scale features, processing through DPSA, and concludes with a final 1×1 convolution, as shown in Figure~\ref{fig:YOLOv10}. DPSA\_SPPF balances feature enhancement with computational efficiency through post concatenation attention application, making it ideal for deployment scenarios demanding both accuracy and inference speed.

\subsection{Focal Generalized IoU Objectness Loss (FGIoU Loss)}
The FGIoU is a loss function designed for a sophisticated composite objective to comprehensively enhance the YOLOv10 detection Framework. This loss function amalgamates three distinct loss components specifically aimed at tackling fundamental issues: class imbalance, inaccurate localization, and poor objectness calibration. FGIoU Loss functions through a well structured pipeline commencing with the TaskAlignedAssigner, which aligns predictions from the dual detection heads of YOLOv10 with corresponding ground truth annotations, subsequently followed by the simultaneous computation of losses and their weighted aggregation. The overall formulation of FGIoU Loss is articulated as: 
\begin{equation}
\mathcal{L}_{\text{FGIoU}} = 7.5 \cdot \mathcal{L}_{\text{GIoU}} + 0.5 \cdot \mathcal{L}_{\text{Focal}} + 1.0 \cdot \mathcal{L}_{\text{ObjFocal}}
\label{eq:fgious_total}
\end{equation}
The loss weights were selected via preliminary experiments and remain stable under small ($\pm$20\%) variations.
Where coefficients are meticulously calibrated to prioritize the precision of bounding boxes while simultaneously balancing contributions from classification and objectness.
\begin{itemize}
    \item Generalized IoU Loss refines the process of bounding box regression by imposing penalties for both insufficient overlap and spatial separation:
    \begin{equation}
    \mathcal{L}_{\text{GIoU}} = 1 - \left( \text{IoU} - \frac{|C \setminus (A \cup B)|}{|C|} \right)
     \label{eq:giou}
    \end{equation}
     Where A and B are predicted and ground truth boxes and c is their minimal enclosing convex hull.
     \item Focal Loss addresses the imbalance between foreground and background by concentrating the training process on challenging examples: 
     \begin{equation}
      \mathcal{L}_{\text{Focal}}(p_t) = -\alpha (1 - p_t)^\gamma \log(p_t)
      \label{eq:focal}
      \end{equation}
      Here, \(p_t\) is the predicted class probability, \(\alpha\) balances class weight, and \(\gamma\) controls the focusing intensity.
      \item Objectness Focal Loss enhances the calibration of confidence through the application of focal weighted binary cross entropy:
      \begin{equation}
       \mathcal{L}_{\text{ObjFocal}}(p_t) = \alpha (1 - p_t)^\gamma \cdot \text{BCE}(p_t, t)
       \label{eq:objfocal}
       \end{equation}

       where \(p_t\) is the objectness score and \(t \in \{0,1\}\) is the ground-truth label.
\end{itemize}

\subsection{Model Evaluation Metrics}
In this experiment, five metrics are employed to evaluate the detection performance of the proposed YOLOv10 model in the context of underwater object detection. The respective formulas are articulated in equations (\ref{eq:precision})-(~\ref{eq:f1}), encompassing Precision, Recall, Average Precision (AP), Mean Average Precision (mAP), and F1 score. 
\begin{equation}
\label{eq:precision}
\text{Precision} = \frac{TP}{TP + FP} \times 100\% 
\end{equation}

\begin{equation}
\label{eq:recall}
\text{Recall} = \frac{TP}{TP + FN} \times 100\% 
\end{equation}

\begin{equation}
\label{eq:ap}
AP = \int_{0}^{1} P(R) \, dR
\end{equation}

\begin{equation}
\label{eq:map}
mAP = \frac{\sum_{j=1}^{c} (AP)_j}{c}
\end{equation}

\begin{equation}
\label{eq:f1}
F1\text{-score} = 2 \times \frac{\text{Precision} \times \text{Recall}}{\text{Precision} + \text{Recall}}
\end{equation}

Where TP denotes true positives, FP signifies false positives, and FN indicates false negatives. Precision measures the model's capability to differentiate negative samples, with higher values indicating superior accuracy in positive predictions. Higher recall denotes a greater proportion of correctly identified positive samples by the model. The F1 score integrates both metrics, with elevated values reflecting a more robust model. The variable C in Equation (\ref{eq:ap}) represents the number of classes evaluated, which was four in this study. For mAP50 calculation, and IoU threshold of 0.5 indicates that model predictions are deemed correct when the intersection over union exceeds 0.5. The F1 score serves to provide a holistic assessment of model performance in instances of significant disparity between accuracy and recall.
\subsection{Experimental Setup}
Experimental procedures were executed utilizing PyTorch 2.x on the Kaggle platform, augmented by GPU acceleration via (Tesla P100). All models were trained for 100 epochs utilizing a $640 \times 640$ input resolution and a batch size of 16. The AdamW optimizer was employed with an initial learning rate of 0.01, scheduled using cosine decay, a momentum coefficient of 0.937, and a weight decay of 0.0005. Furthermore, early stopping was instituted with a patience of 10 epochs to alleviate the risk of overfitting. The model achieves an average inference time of 2.1 ms per image (approximately 476 FPS) at 640×640 resolution.  

\section{Results and Discussion}
The section presents the experimental evaluation of model enhancements, beginning with an ablation study conducted on the baseline YOLOv10n~\cite{yolov10} model to assess the contributions of the DPSA\_SPPF module and FGIoU loss. The proposed method is compared with existing YOLO variants on the RUOD and DUO datasets, highlighting it's impact on detection accuracy, precision, and robustness in complex underwater environments. Due to space limitations, only overall metrics are reported, while consistent PR trends are observed.

\begin{table}[h]
\centering
\caption{Experimental comparison with the baseline model.}
\label{tab:comparison}
\scriptsize
\setlength{\tabcolsep}{4pt}
\begin{tabular}{lccccc}
\hline
Dataset & Method & DPSA\_SPPF & FGIoU Loss & mAP50 (\%) & mAP50:95 (\%) \\
\hline
RUOD & YOLOv10n & --  & --  & 82.2 & 58.8 \\
     & YOLOv10n & Yes & --  & 88.3 (+6.1) & 65.1 (+6.3) \\
     & YOLOv10n & --  & Yes & 88.0 (+5.8) & 64.9 (+6.1) \\
     & YOLOv10n & Yes & Yes & 88.9 (+6.7) & 66.5 (+7.7) \\
\hline
DUO  & YOLOv10n & --  & --  & 81.8 & 63.5 \\
     & YOLOv10n & Yes & --  & 85.7 (+3.9) & 66.4 (+2.9) \\
     & YOLOv10n & --  & Yes & 87.4 (+5.6) & 68.0 (+4.5) \\
     & YOLOv10n & Yes & Yes & 88.0 (+6.2) & 69.1 (+5.6) \\
\hline
\end{tabular}
\end{table}

\begin{table*}[t]
\centering
\caption{Comparison of the improved YOLOv10n variants with state-of-the-art methods for underwater object detection on the RUOD and DUO datasets}
\label{tab:uwdd_duo_comparison}
\resizebox{\textwidth}{!}{%
\begin{tabular}{l cccc cccc c}
\hline
Model 
& \multicolumn{4}{c}{RUOD Dataset} 
& \multicolumn{4}{c}{DUO Dataset} 
& Parameters \\
\cline{2-5} \cline{6-9}
& Prec. \% & Rec. \% & mAP@0.5 \% & mAP@0.5--0.95 \%
& Prec. \% & Rec. \% & mAP@0.5 \% & mAP@0.5--0.95 \% \\
\hline
YOLOv8n 
& 83.8 & 75.0 & 82.6 & 58.4
& 81.9 & 75.7 & 82.7 & 64.1
& 3.1M \\
\hline
YOLOv8s 
& 85.7 & 78.5 & 85.6 & 61.4
& 86.5 & 76.7 & 83.4 & 67.0
& 11.1M \\
\hline
YOLOv8m 
& 85.6 & 77.9 & 85.8 & 62.1
& 85.7 & 76.8 & 83.4 & 68.4
& 25.8M \\
\hline
YOLOv9t 
& 84.1 & 75.1 & 82.4 & 59.0
& 86.5 & 77.6 & 85.5 & 66.4
& 2.01M \\
\hline
YOLOv10n 
& 82.3 & 76.0 & 82.2 & 58.8
& 88.3 & 68.4 & 81.8 & 63.5
& 2.7M \\
\hline
YOLOv10s 
& 84.4 & 78.2 & 84.3 & 61.9
& 84.6 & 78.5 & 86.1 & 66.3
& 8.1M \\
\hline
YOLOv11n 
& 83.2 & 76.5 & 82.8 & 58.8
& 84.0 & 77.6 & 85.8 & 66.7
& 2.60M \\
\hline
Dynamic YOLO \cite{P10}
& -- & -- & -- & --
& -- & -- & 86.7 & 68.6
& 8.2M \\
\hline
DPSA\_YOLOv10n (Ours) 
& 85.7 & 80.7 & 88.3 & 65.1
& 84.5 & 78.2 & 85.7 & 66.4
& 2.728M \\
\hline
FGIoU\_YOLOv10n (Ours) 
& 85.7 & 80.5 & 88.0 & 64.9
& 87.1 & 78.0 & 87.4 & 68.0
& 2.77M \\
\hline
DPSA\_FGIoU\_YOLOv10n (Ours) 
& 86.7 & 82.1 & 88.9 & 66.5
& 87.2 & 78.6 & 88.0 & 69.1
& 2.8M \\
\hline
\end{tabular}
}
\end{table*}

\subsection{Component Wise Impact Analysis of Model Enhancements}
Table~\ref{tab:comparison} presents a component-wise ablation analysis evaluating the individual and combined effects of the DPSA\_SPPF module and the FGIoU loss on the baseline YOLOv10n model across the RUOD and DUO datasets. On the RUOD dataset, the baseline model achieves an mAP@0.5 of 82.2\% and an mAP@0.5:0.95 of 58.8\%. Introducing the DPSA\_SPPF module alone improves feature representation, increasing mAP@0.5 to 88.3\% (+6.1\%) and mAP@0.5:0.95 to 65.1\% (+6.3\%). Replacing the conventional loss function with FGIoU enhances localization accuracy, resulting in mAP@0.5 and mAP@0.5:0.95 scores of 88.0\% (+5.8\%) and 64.9\% (+6.1\%), respectively. The joint application of both enhancements yeilds optimal performance, achieving an mAP@0.5 of 88.9\% and an mAP@0.5:0.95 of 66.5\%. Compared with the baseline, this corresponds to gains of +6.7\% and +7.7\%, respectively. Consistent trends are observed on the DUO dataset, where the baseline mAP@0.5 and mAP@0.5:0.95 scores of 81.8\% and 63.5\% are improved to 85.7\% (+3.9\%) and 66.4\% (+2.9\%) with DPSA\_SPPF, and to 87.4\% (+5.6\%) and 68.0\% (+4.5\%) with FGIoU. The combined approach achieves optimal performance with mAP@0.5 and mAP@0.5:0.95 scores of 88.0\% and 69.1\%, indicating enhancements of +6.2\% and +5.6\% over the baseline. These results demonstrate that the proposed architectural attention enhancement and loss optimization provide complementary benefits, yielding consistent and significant performance improvements across both datasets while preserving the lightweight nature of the baseline model.

\subsection{Comparative Evaluation with State-of-the-Art Models}
Table~\ref{tab:uwdd_duo_comparison} compares the proposed DPSA\_FGIoU\_YOLOv10n model in comparison to leading YOLO variants, which encompass YOLOv8 (n, s, m), YOLO9t, YOLOv10 (n, s), and YOLOv11n, utilizing the RUOD and DUO datasets. Within the RUOD dataset, the proposed model shows the hightest level of detection efficacy, achieving 86.7\% precision, 82.1\% recall, 88.9\% mAP@0.5 (±0.3 over three runs), 66.5\% mAP@0.5-0.95, thereby surpassing YOLOv10n (82.3\% precision, 76.0\% recall, 82.2\% mAP@0.5) and the YOLOv8 variants, which exhibit a lower overall mAP despite their competitive precision metrics. Conversely, on the DUO dataset, although YOLOv10n displays marginally elevated 88.3\% precision, the proposed model secures superior 78.6\% recall and mAP metrics (88.0\% mAP@0.5 (±0.3 over three runs), 69.1\% mAP@0.5-0.95), thereby showing a more balanced and reliable capability for object detection in challenging underwater environments. Notably, the proposed model maintains a lightweight architecture with only 2.8M parameters, which is substantially smaller than YOLOv8s and YOLOv8m, demonstrating an effective trade-off between detection performance and computational efficiency. These results highlight the superiority of DPSA\_FGIoU\_YOLOv10n for real-time underwater object detection under resource-constrained conditions.

\begin{figure}[htbp]
    \centering
    \includegraphics[width=.90\linewidth, height=.30\textheight]{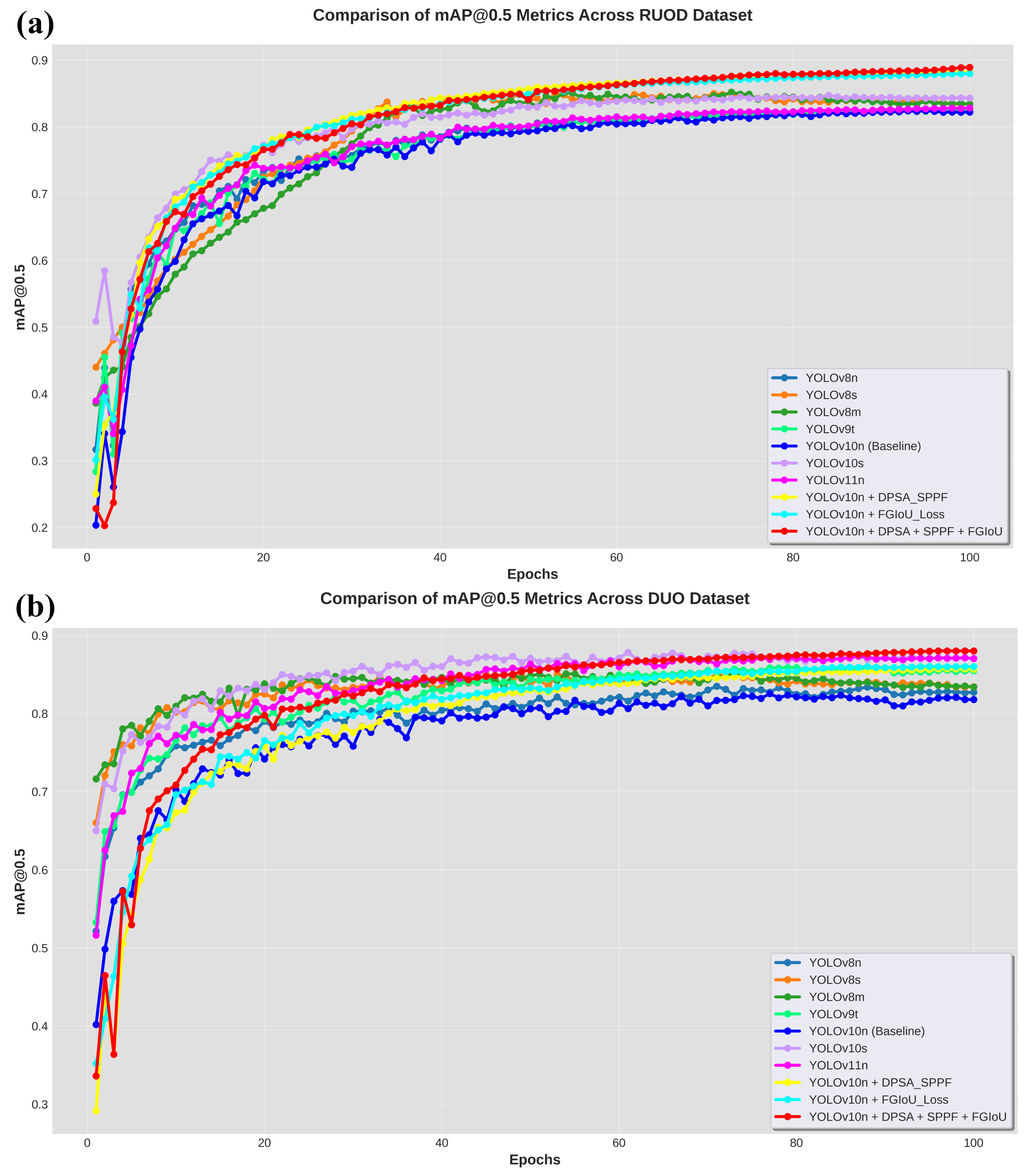}
    \caption{Comparison of mAP@0.5 Metrics Across (a) RUOD, and (b) DUO dataset.}
    \label{fig:R_C}
\end{figure}

\subsection{In Depth Analysis of Model Performance Across Datasets}
Figure~\ref{fig:R_C} presents a detailed comparison of mean Average Precision (mAP@.5) metrics for the RUOD and DUO datasets. The plots demonstrate the performance trajectories of various YOLO variants across epochs, particularly highlighting the proposed DPSA\_FGIoU\_YOLOv10n model. The DPSA\_FGIoU\_YOLOv10n model consistently surpasses the baseline YOLOv10n in mAP@0.5 across both datasets, with improved values at every epoch. The incorporation of the DPSA\_SPPF attention module and FGIoU loss function significantly enhances model performance, faciliating a continuous increase in mAP scores during training. Conversely, although YOLOv8 variants exhibit competitive precision, they are less consistent and overall inferior to the DPSA\_FGIoU\_YOLOv10n model. The findings highlight the benefits of the proposed architectural enhancements, showcasing the model's capacity for high detection accuracy and strong performance, thus rendering it suitable for practical applications necessitating precision and computational efficiency.

%model with existing YOLO variants across the UWD and DUO datasets. While YOLOv10 variants performed competitively, the DPSA\_FGIoU model consistently outperformed them, achieving 86.7\% precision, 82.1\% recall, 88.9\% mAP50, and 66.5\% mAP50:95 on UWD dataset and 87.2\% precision, 78.6\% recall, 88.0\% mAP50, and 69.1\% mAP50:95 on DUO dataset. The results underscore the model's robustness in diverse and challenging underwater conditions while maintaining an efficient parameter count of 2.8M, suggesting favorable deployment potential. Collectively, the enhancements effectively bolster feature extraction and localization, validating the model's state-of-the-art performance and practical applicability.

\section{Conclusion}
This paper presented an efficient and lightweight underwater object detection framework built upon YOLOv10 to address challenges arising from underwater visual degradation and small object detection. By incorporating a deterministic Multi-Stage Adaptive Enhancement preprocessing pipeline, the Dual-Pooling Sequential Attention mechanism and FGIoU loss enhance feature representation, localization accuracy, and objectness calibration while maintaining computational efficiency. Experimental findings on the RUOD and DUO datasets reveal substantial and consistent advancements compared to the baseline YOLOv10n, achieving absolute mAP@0.5 gains of up to 6.7\% and 6.2\%, respectively, while preserving a compact architecture with only 2.8M parameters. These improvements stem from the complementary effects of attention based feature refinement and loss level optimization. Overall, the proposed framework provides a practical and reliable solution for real-time underwater perception on embedded platforms and also explore temporal feature modeling and domain adaptation to further enhance robustness in dynamic underwater environments.

%This work proposes a lightweight and robust underwater object detection framework that enhances YOLOv10 by integrating deterministic image enhancement, attention-guided multi-scale feature refinement, and an optimized loss function. The MAE UVP module effectively mitigates underwater visual degradation by restoring color fidelity, enhancing contrast, suppressing haze, and preserving structural details without introducing additional learnable parameters. In contrast, the DPSA SPPF module strengthens feature representation and suppresses background noise with minimal computational overhead. The proposed Focal Generalized IoU-based objectness loss further improves localization accuracy, class-imbalance handling, and confidence calibration, resulting in stable optimization and improved detection performance. Extensive evaluations on the RUOD and DUO datasets demonstrate consistent improvements over recent YOLO variants in terms of precision, recall, and mean Average Precision, while maintaining a compact model size suitable for real-time deployment. Future work will focus on extending the framework to video-based underwater detection, improving robustness under extreme visibility conditions, and validating the approach on larger and more diverse underwater datasets to further enhance generalization and practical applicability.

\bibliographystyle{IEEEtran}
\bibliography{Ref}
\end{document}